\documentclass{article}

\usepackage[preprint]{neurips_2026}

\usepackage[utf8]{inputenc} 
\usepackage[T1]{fontenc}    
\usepackage{hyperref}       
\hypersetup{
    colorlinks=true,
    linkcolor=black,
    citecolor=black,
    urlcolor=blue!60!black
}
\usepackage{url}            
\usepackage{booktabs}       
\usepackage{amsfonts}       
\usepackage{amsmath,amssymb}
\usepackage{nicefrac}       
\usepackage{microtype}      
\usepackage{xcolor}         
\usepackage{graphicx}
\usepackage{enumitem}

\title{From Edges to Depth: Probing the Spatial Hierarchy in Vision Transformers}

\author{%
  Jainum Sanghavi \\
  Northeastern University\\
  \texttt{sanghavi.j@northeastern.edu} \\
}

\begin{document}

\maketitle

\begin{abstract}
Vision Transformers trained only on image classification routinely transfer to tasks that demand spatial understanding, yet they receive no spatial supervision during pretraining.  We ask \emph{where} and \emph{how robustly} such structure is encoded.  Probing a frozen ViT-B/16 layerwise for two complementary properties, local patch boundaries (BSDS500) and per-patch depth (NYU Depth V2), reveals a clear hierarchy: boundary structure becomes linearly decodable at layers 5--6 (AP = 0.833), while depth, which requires integrating global cues, peaks two to three layers later at layer 8 (MAE = 0.0875).  Both signals collapse at the final classification layer, and random-weight controls confirm the encodings are learned rather than architectural.  Causal interventions add specificity: ablating the single direction a linear depth probe reads degrades depth decoding by up to 165\%, while ablating any other direction changes it by less than 1\%.  Targeted activation patching along that direction shows the depth signal is partially re-derived at each layer rather than passively carried in the residual stream, with mid-layer interventions persisting most strongly downstream.  The result is that a classification-trained ViT develops an actively maintained spatial hierarchy that mirrors the early-to-late progression observed in the primate visual cortex.
\end{abstract}

\section{Introduction}

Vision Transformers~\citep{dosovitskiy2021vit} trained on large-scale image classification achieve strong performance on tasks that implicitly require spatial understanding, including detection, segmentation, and depth estimation, yet they never receive explicit spatial supervision.  A ViT trained on ImageNet sees only a class label per image.  This raises a fundamental interpretability question: \emph{what structural and spatial information is implicitly encoded in ViT representations, and where does it emerge?}

Prior work has shown that ViT representations contain spatially meaningful structure.  \citet{raghu2021vision} demonstrated that ViTs incorporate global information earlier than CNNs, using CKA similarity and linear probes on ImageNet classes.  \citet{caron2021dino} showed that self-supervised ViTs develop object-segmenting attention maps.  \citet{zhou2024elvit} observed that same-object patches cluster in representation space across layers.  However, these studies either probe for the model's own training objective, report results at a single layer, or stop at correlational evidence.  None systematically resolves \emph{which} spatial properties emerge at \emph{which} layers, controls for architectural confounds, or validates findings causally.

We address this with \textbf{probing classifiers}~\citep{alain2017understanding}: lightweight models trained on frozen hidden states to predict a target property.  If a linear probe succeeds, the information is \emph{linearly accessible}, meaning it is encoded in a separable subspace.  We conduct two parallel probing experiments on the same ViT-B/16:

\begin{enumerate}[noitemsep,topsep=2pt,leftmargin=*]
    \item \textbf{Boundary detection} (BSDS500): binary classification of whether each $16\!\times\!16$ patch overlaps a ground-truth contour.  This probes \emph{local structural} sensitivity, specifically the model's ability to distinguish patches at intensity or semantic discontinuities from interior patches.
    \item \textbf{Depth regression} (NYU Depth V2): predicting the mean depth of each patch in normalised units.  This probes \emph{spatial layout}, an inherently 3D property that requires integrating global cues like perspective, occlusion, and relative object size.
\end{enumerate}

However, probing is correlational~\citep{hewitt2019designing}: it shows information is \emph{present} but not that the model \emph{uses} it.  We therefore add a causal layer: (a)~probe-direction ablation, which erases the depth-encoding direction and measures the impact, and (b)~targeted activation patching, which replaces only the depth-direction component at a given layer and tracks downstream persistence.  This targeted, single-direction patching is a sharper intervention than standard full-layer patching~\citep{heimersheim2024patching}, because it isolates one specific encoding axis rather than replacing the entire representation.

Our contributions:
\begin{enumerate}[noitemsep,topsep=2pt,leftmargin=*]
    \item A layerwise probing study of a classification-only ViT that jointly resolves two complementary spatial properties, boundaries and depth, with matched random-weight controls, revealing a cross-task spatial hierarchy (boundary peaks at L5--6, depth at L8).
    \item Evidence that both properties are linearly accessible: linear probes match or outperform MLP probes with $257\times$ more parameters, and the depth signal is concentrated in a single linear direction (165\% MAE increase upon ablation vs.\ $<$1\% for random directions).
    \item Causal validation via targeted activation patching showing the depth signal is actively maintained rather than passively carried: mid-layer interventions retain 76\% of the effect one layer downstream, while early-layer interventions decay to $<$5\% within two layers.
\end{enumerate}

Code and data are available at \url{https://github.com/JainumSanghavi/ProbingViTs}.

\section{Related work}

\textbf{Probing methodology.}  \citet{alain2017understanding} introduced linear probes on frozen representations to measure linearly accessible information at each layer.  \citet{hewitt2019structural} extended this with structural probes that recover syntax trees from BERT's geometry, motivating probing for structured spatial properties rather than simple categorical attributes.  \citet{hewitt2019designing} showed probes can recover information the model does not use, making results correlational and motivating control tasks.  \citet{lipton2018mythos} raises a broader caution against underspecified interpretability claims, warning that many papers assert interpretability without operationally defining the term.  \citet{heimersheim2024patching} provide guidance on activation patching as a causal complement to probing.

\textbf{ViT interpretability.}  \citet{raghu2021vision} compared ViT and CNN representations using CKA similarity and linear probes on ImageNet classes, finding ViTs incorporate global information earlier than CNNs, but did not probe for external spatial properties or include causal interventions.  \citet{caron2021dino} showed self-supervised ViTs develop object-segmenting attention maps.  \citet{chowdhury2025promptcam} showed spatially localised information is recoverable from frozen ViT patch embeddings, and \citet{zhou2024elvit} showed same-object patches cluster in representation space across layers, a qualitative observation consistent with our quantitative boundary probing results.  \citet{qiang2025iavit} showed interpretability can be trained into a ViT by modifying the training objective, a complementary approach to our post-hoc probing methodology.

\textbf{The gap.}  Existing work establishes that ViTs encode structured information, that probing is correlational, and that activation patching can make it causal.  To our knowledge, no prior study combines a layerwise analysis that (a) targets more than one spatial property to reveal their relative ordering, (b) uses random-weight controls to separate learned from architectural effects, and (c) follows up with targeted causal interventions on the probe's own encoding direction.  Our work combines all three.

\section{Methodology}

\subsection{Model and feature extraction}

We probe \texttt{google/vit-base-patch16-224-in21k} (ViT-B/16, 86M parameters).  Each $224\!\times\!224$ image is divided into 196 non-overlapping $16\!\times\!16$ patches, projected into a 768-dimensional token space, and processed through 12 transformer blocks.  We extract hidden states at all 13 layers (embedding layer plus 12 transformer blocks), yielding a tensor of shape $(13, 196, 768)$ per image.  The CLS token is excluded because our analysis targets per-patch spatial properties.  We extract hidden states for both the \textbf{pretrained} and a \textbf{randomly initialised} ViT-B/16.  The random model serves as a control: any information a probe recovers from random weights reflects dataset statistics or positional encoding artefacts rather than learned representations.

\subsection{Datasets and labels}

\begin{table}[t]
\centering
\caption{Dataset summary.}
\label{tab:datasets}
\begin{tabular}{@{}llrrr@{}}
\toprule
Task & Labels & Train & Val & Test \\
\midrule
Boundary (BSDS500) & Binary/patch & 200 & 100 & 200 \\
Depth (NYU V2)     & Mean depth/patch   & 675 & 120 & 654 \\
\bottomrule
\end{tabular}
\end{table}

\textbf{Boundary labels.}  For each $16\!\times\!16$ patch, we check whether it overlaps any annotator-agreed boundary contour in the BSDS500 ground truth.  BSDS500 provides multiple human annotations per image (typically 4 to 9 annotators), and we use the consensus boundaries where a majority of annotators agree.  Labels are binary: 1 (boundary) or 0 (non-boundary).  Boundary detection is a local structural property: it depends on whether adjacent patches exhibit contrast in intensity or semantics, not on where in the image the patch is positioned.

\textbf{Depth labels.}  Depth maps from NYU Depth V2~\citep{silberman2012nyudepth} are resized to $224\!\times\!224$ via bilinear interpolation, then the mean depth within each $16\!\times\!16$ patch is computed.  Values are normalised by dividing by 10.0\,m (the approximate dataset maximum), mapping all labels to $[0,1]$.  Depth is a spatial property that requires integrating global cues: context from surrounding patches (furniture scale, wall convergence, floor plane orientation) determines a patch's depth, not its spatial position alone.

\subsection{Probe architectures and training}

\begin{table}[t]
\centering
\caption{Probe configurations.  Both tasks use Adam (lr$=10^{-3}$, weight decay $10^{-4}$), batch size 512, max 100 epochs, patience 10.}
\label{tab:probe_configs}
\begin{tabular}{@{}lll@{}}
\toprule
 & Boundary & Depth \\
\midrule
Linear probe & $768 \to 1$ (769 params) & $768 \to 1$ (769 params) \\
MLP probe & $768 \to 256 \to 1$ (197K params) & $768 \to 256 \to 1$ (197K params) \\
Loss & BCEWithLogits & MSE \\
Primary metric & AP (PR-AUC) & MAE \\
Early stopping on & val F1 (max) & val MAE (min) \\
\bottomrule
\end{tabular}
\end{table}

Each experiment produces $13 \times 2 \times 2 = 52$ probe runs (13 layers $\times$ 2 probe types $\times$ 2 model initialisations).  All probes are applied independently per patch.

The choice of boundary detection and depth regression as the two probing targets is deliberate.  They differ along the axis that matters most for understanding how a transformer builds spatial representations: \emph{locality}.  Boundary detection is a local structural property: it depends only on contrast between a patch and its immediate neighbours, and self-attention can aggregate this within a few layers.  Depth is a global spatial property: inferring how far away a surface is requires integrating perspective cues, occlusion relationships, and relative object sizes from across the full image.  If these two properties peak at the same layer, locality does not matter for representational emergence.  If they peak at different layers, the offset directly measures how many additional layers of cross-patch communication the model needs to transition from local structure to global spatial layout.

We report \textbf{Average Precision (AP)}, the area under the precision-recall curve, as the primary metric for boundary detection.  AP is threshold-free and provides a single continuous summary of the probe's ranking quality across all decision thresholds.  The threshold-free property is useful here because boundary detection is imbalanced: across the dataset, 32\% of patches overlap a consensus boundary contour (29\% in validation, 34\% in test), with the remaining patches being interior.  This imbalance arises because we label a patch as positive if \emph{any} of its 256 pixels lies on a consensus contour; a stricter threshold would reduce the positive rate further.  A majority-class classifier would achieve high accuracy while being uninformative; AP penalises such classifiers because their precision-recall curve collapses.  We also report F1 at threshold 0.5 for comparability with prior work, and use val-F1 as the early stopping target.

\subsection{Causal interventions}

\textbf{Probe direction ablation.}  The linear probe's weight vector $\mathbf{w} \in \mathbb{R}^{768}$ defines the ``depth direction,'' the single axis the probe reads.  We project it out of every hidden state:
\begin{equation}
\mathbf{h}' = \mathbf{h} - (\mathbf{h} \cdot \hat{\mathbf{w}})\hat{\mathbf{w}}
\end{equation}
where $\hat{\mathbf{w}} = \mathbf{w}/\|\mathbf{w}\|$.  The critical control: we repeat this with 10 random directions.  If ablating the probe direction causes far more damage than ablating any random direction, the depth signal is concentrated in a specific direction rather than diffusely distributed.  We also run a dose-response experiment, scaling the ablation by $\alpha \in \{0.0, 0.1, \ldots, 1.0\}$.

\textbf{Targeted activation patching.}  Standard activation patching~\citep{heimersheim2024patching} typically replaces an entire layer's activations or a single component (attention head, MLP block).  We apply a more targeted variant: for 20 image pairs with maximal depth contrast, we replace \emph{only the depth-direction component} at layer $L$:
\begin{equation}
\mathbf{h}'_\text{src} = \mathbf{h}_\text{src} - (\mathbf{h}_\text{src} \cdot \hat{\mathbf{w}}_L)\hat{\mathbf{w}}_L + (\mathbf{h}_\text{dst} \cdot \hat{\mathbf{w}}_L)\hat{\mathbf{w}}_L
\end{equation}
where $\hat{\mathbf{w}}_L$ is the probe's weight direction at layer $L$.  The other 767 dimensions remain untouched.  We then measure how much the depth probe's predictions at downstream layers $T \geq L$ shift toward the destination's depth.  This distinguishes two hypotheses: (a) depth is passively carried in the residual stream (effect persists near 1.0), or (b) depth is partially re-derived at each layer from the full representation (effect decays).

\section{Boundary detection results}

\begin{figure}[t]
\centering
\includegraphics[width=0.8\linewidth]{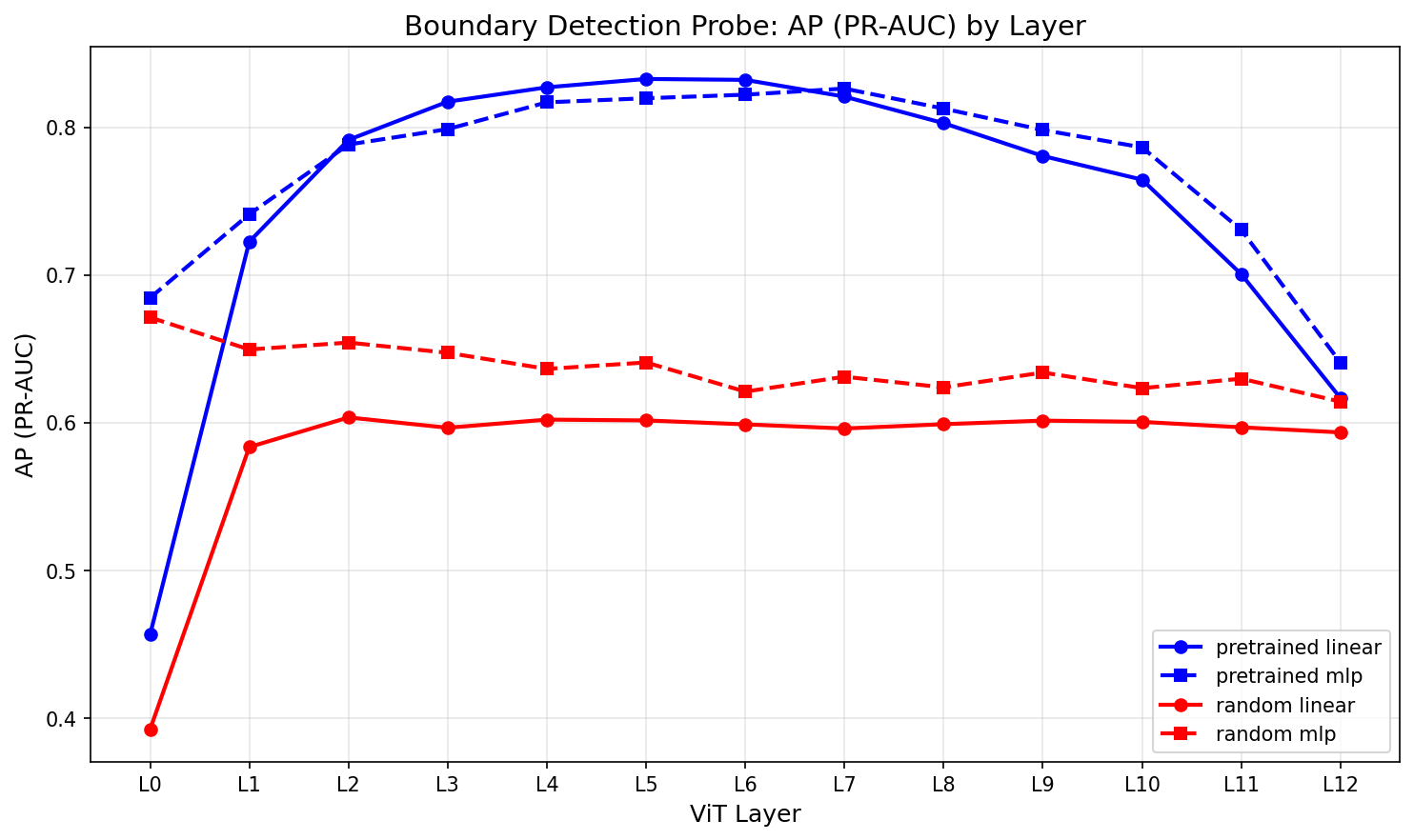}
\caption{Boundary detection AP by layer.  Pretrained ViT (blue) peaks at layer 5 (AP = 0.833); random ViT (red) stays flat around 0.60.  The $+$0.23 AP gap is the learned signal.}
\label{fig:boundary_ap}
\end{figure}

The pretrained linear probe achieves \textbf{peak AP = 0.833 at layer 5} (F1 = 0.747, accuracy 79.3\%), rising steeply from AP = 0.456 at layer 0 and declining to 0.617 at layer 12.  The MLP probe peaks at AP = 0.826 (layers 6--7, accuracy 83.1\%).  The random-weight baseline is flat across all layers (AP $\approx$ 0.60), confirming that boundary encoding is \emph{learned}.

\begin{table}[t]
\centering
\caption{Boundary detection at selected layers.  The pretrained-to-random AP gap peaks at $+$0.23 at layers 5--6.}
\label{tab:boundary_results}
\begin{tabular}{@{}crrrrr|rr@{}}
\toprule
 & \multicolumn{5}{c|}{Linear probe (pretrained)} & \multicolumn{2}{c}{Rand.} \\
Layer & AP & F1 & Acc & Prec & Rec & AP & F1 \\
\midrule
0  & 0.456 & 0.527 & 0.445 & 0.369 & 0.917 & 0.392 & 0.503 \\
3  & 0.817 & 0.739 & 0.786 & 0.626 & 0.901 & 0.597 & 0.592 \\
5  & \textbf{0.833} & 0.747 & 0.793 & 0.635 & 0.906 & 0.601 & 0.611 \\
6  & 0.832 & \textbf{0.747} & \textbf{0.797} & \textbf{0.642} & 0.894 & 0.599 & 0.594 \\
8  & 0.803 & 0.720 & 0.770 & 0.610 & 0.881 & 0.599 & 0.597 \\
12 & 0.617 & 0.612 & 0.663 & 0.500 & 0.790 & 0.593 & 0.591 \\
\bottomrule
\end{tabular}
\end{table}

At layer 0, the high recall (0.917) paired with low precision (0.369) reflects the probe predicting ``boundary'' for most patches: before any attention, patch embeddings lack the contextual information needed to suppress false positives.  As layers progress, precision climbs steadily (reaching 0.642 at layer 6) while recall remains high, indicating the model develops increasingly selective boundary representations.

The MLP probe peaks slightly lower than the linear probe on AP (0.826 vs.\ 0.833) but achieves higher accuracy (83.1\% vs.\ 79.7\%) due to a better precision-recall tradeoff at the fixed 0.5 threshold.  This marginal MLP advantage vanishes when evaluated threshold-free, suggesting that for boundary detection, the linear probe already captures the relevant subspace and the MLP's extra capacity provides negligible benefit.  The close match between linear and MLP probes on both tasks strengthens the claim that the encoded information is linearly separable.

The rate of improvement is worth examining.  Boundary AP rises from 0.456 at layer 0 to 0.817 at layer 3, a gain of 0.361 in just three transformer blocks, then plateaus between layers 3 and 6 with only a 0.016 additional gain.  This steep early rise indicates that the first few self-attention layers are where the bulk of boundary-relevant computation occurs, with subsequent layers refining rather than fundamentally changing the representation.

\section{Depth probing results}

\begin{figure}[t]
\centering
\includegraphics[width=0.8\linewidth]{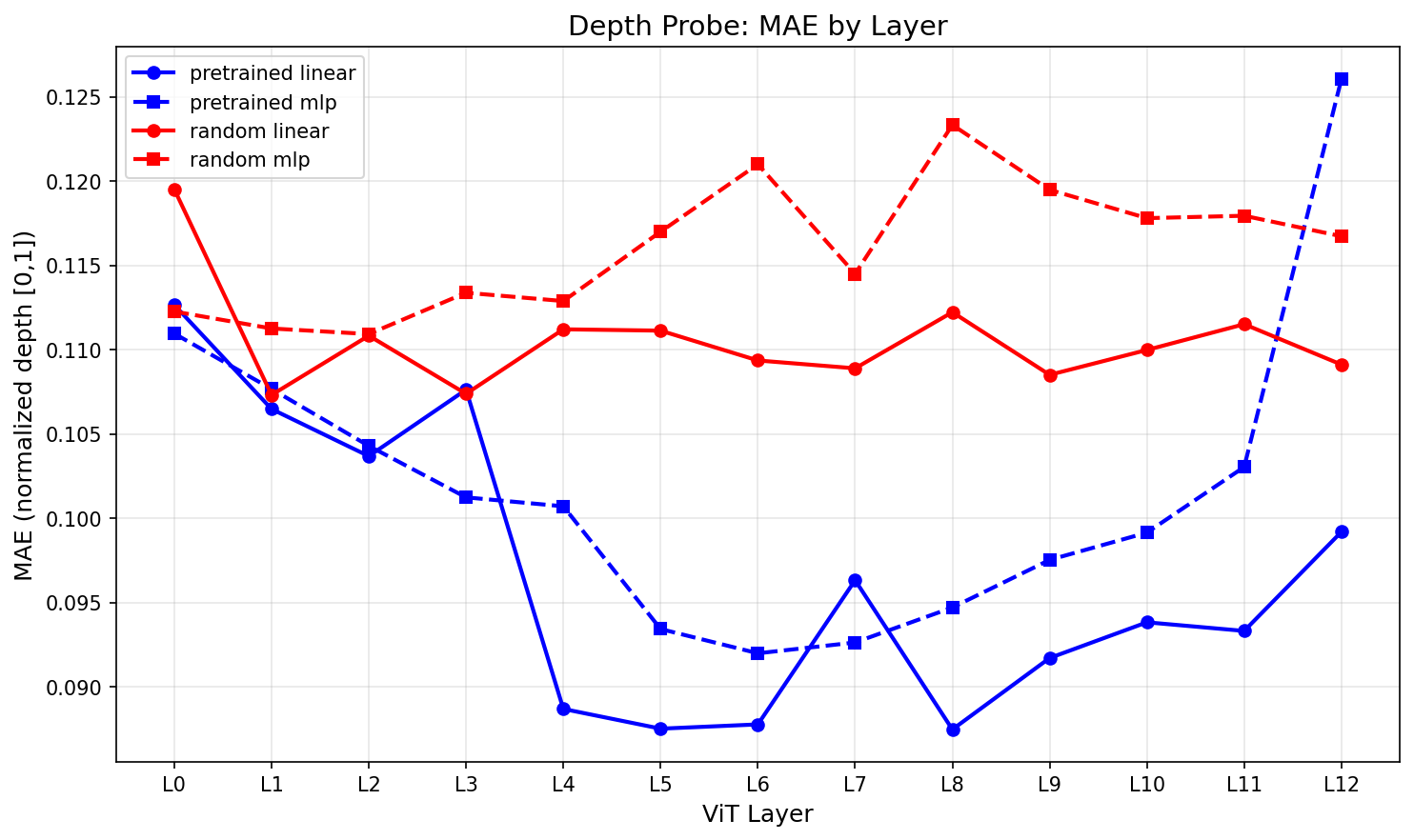}
\caption{Depth probing MAE by layer.  Pretrained ViT (blue) reaches best MAE at layers 5--8; random ViT (red) is flat around 0.11.}
\label{fig:depth_mae}
\end{figure}

The pretrained linear probe achieves \textbf{best MAE = 0.0875} (0.875\,m) at \textbf{layer 8}, a 23\% reduction over the random-weight baseline (MAE $\approx$ 0.107).  The MLP probe peaks earlier at layer 6 (MAE = 0.0920) but performs worse on the test set despite having $257\times$ more parameters, indicating depth information is \textbf{primarily linearly accessible}.

\begin{table}[t]
\centering
\caption{Depth probing at selected layers (MAE, normalised; $\times$10 = metres).}
\label{tab:depth_results}
\begin{tabular}{@{}c|rr|rr|rr|rr@{}}
\toprule
 & \multicolumn{4}{c|}{Pretrained} & \multicolumn{4}{c}{Random} \\
Layer & \multicolumn{2}{c|}{Linear} & \multicolumn{2}{c|}{MLP} & \multicolumn{2}{c|}{Linear} & \multicolumn{2}{c}{MLP} \\
 & MAE & RMSE & MAE & RMSE & MAE & RMSE & MAE & RMSE \\
\midrule
0  & .1127 & .1562 & .1110 & .1541 & .1195 & .1662 & .1123 & .1566 \\
4  & .0887 & .1223 & .1007 & .1424 & .1112 & .1562 & .1129 & .1570 \\
5  & .0875 & .1222 & .0934 & .1337 & .1111 & .1566 & .1170 & .1612 \\
6  & .0878 & .1228 & \textbf{.0920} & \textbf{.1317} & .1094 & .1535 & .1210 & .1657 \\
8  & \textbf{.0875} & \textbf{.1220} & .0947 & .1346 & .1122 & .1579 & .1233 & .1672 \\
12 & .0992 & .1344 & .1261 & .1810 & .1091 & .1528 & .1167 & .1609 \\
\bottomrule
\end{tabular}
\end{table}

The linear probe outperforming the MLP is notable.  An MLP with 197K parameters should recover any information a 769-parameter linear probe can.  That it does not indicates the MLP overfits: the depth signal occupies a clean linear subspace, and additional capacity fits noise rather than useful nonlinear structure.  This is confirmed by the ablation results (Section~\ref{sec:causal}), which show a single direction accounts for nearly all of the depth signal.

The MLP degrades sharply at layer 12 (MAE = 0.126 vs.\ 0.099 for the linear probe), suggesting the final layer's class-specific representation is particularly hostile to nonlinear decoders that overfit to its narrow subspace.  The linear probe degrades more gracefully because it has fewer parameters to overfit with and simply reads a single direction that, while less informative at layer 12, still carries some residual depth signal.

The random-weight baseline is worth examining separately.  Its best MAE ($\approx$0.107) is non-trivial, meaningfully better than a constant predictor that always outputs the dataset mean.  This floor of decodability likely reflects positional encoding priors: in NYU Depth V2's indoor scenes, the bottom of the image tends to be closer than the top, and positional encodings allow even a random ViT to capture this spatial regularity.  The pretrained model's 23\% improvement over this floor represents depth features that go beyond simple position-depth correlations, encoding scene-specific cues like furniture scale and surface orientation.

\section{Cross-task comparison: the spatial hierarchy}

\begin{figure}[t]
\centering
\includegraphics[width=0.85\linewidth]{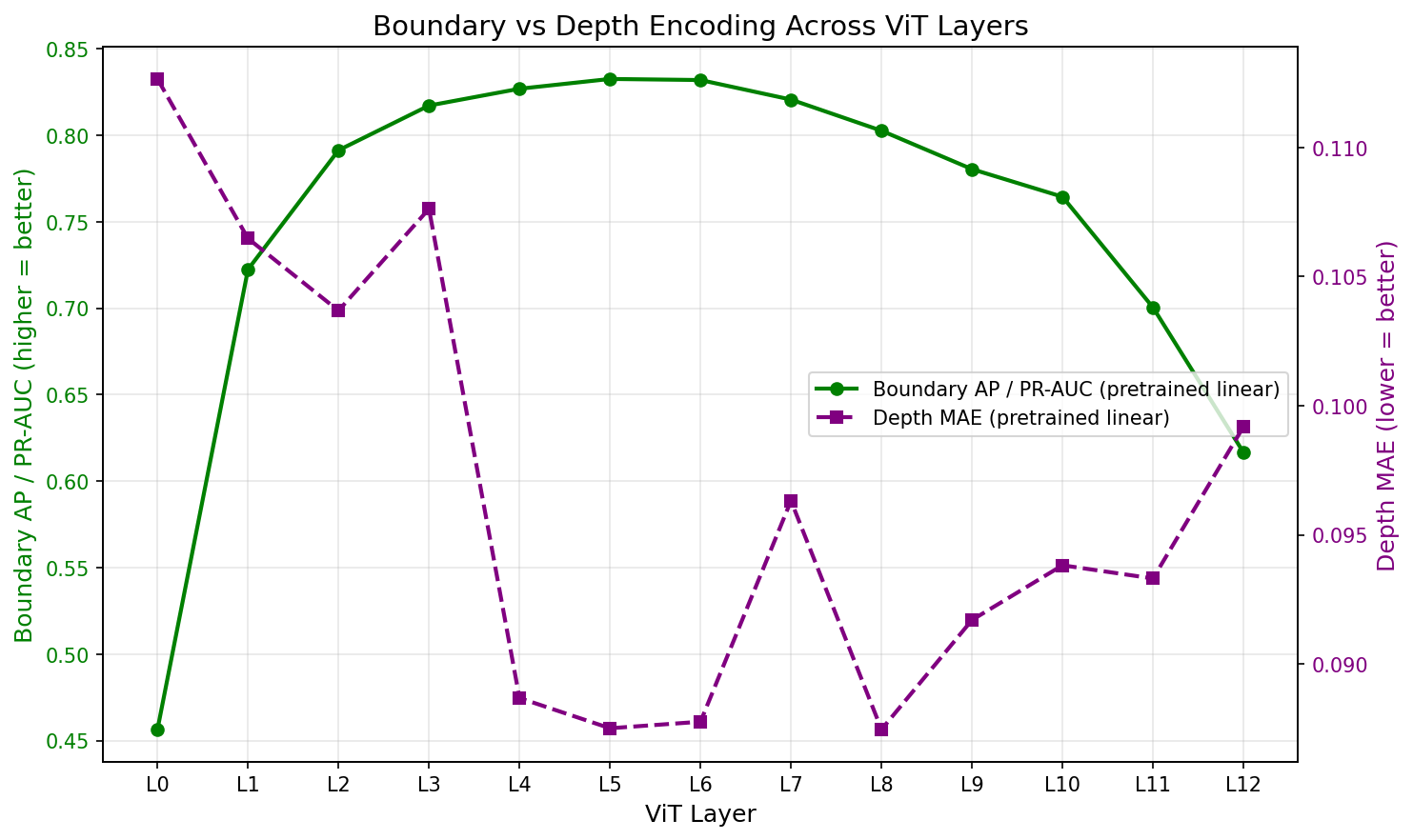}
\caption{Cross-task comparison (pretrained linear probes).  Green/left axis: boundary AP (higher = better).  Purple/right axis: depth MAE (lower = better).  The 2--3 layer offset between their peaks reveals a spatial hierarchy.}
\label{fig:cross_task}
\end{figure}

The 2--3 layer offset between the boundary peak (L5, AP = 0.833) and the depth peak (L8, MAE = 0.0875) is the central finding of this work.  It is consistent across both linear and MLP probes and reveals a \textbf{spatial hierarchy} within the transformer stack:

\begin{table}[t]
\centering
\caption{Cross-task comparison at selected layers (pretrained linear probes).}
\label{tab:cross_task}
\begin{tabular}{@{}crrrr@{}}
\toprule
Layer & Bdry F1 & Bdry AP & Depth MAE & Depth RMSE \\
\midrule
0  & 0.527 & 0.456 & 0.1127 & 0.1562 \\
3  & 0.739 & 0.817 & 0.1076 & 0.1485 \\
5  & 0.747 & \textbf{0.833} & 0.0875 & 0.1222 \\
6  & \textbf{0.747} & 0.832 & 0.0878 & 0.1228 \\
8  & 0.720 & 0.803 & \textbf{0.0875} & \textbf{0.1220} \\
12 & 0.612 & 0.617 & 0.0992 & 0.1344 \\
\bottomrule
\end{tabular}
\end{table}

\textbf{Structural features emerge first.}  Boundary detection depends on local contrast between adjacent patches, which self-attention can aggregate within a few layers.  By layer 5, the model has enough local context to distinguish edge patches from interior patches at near-peak accuracy.

\textbf{Spatial features require more computation.}  Depth estimation requires integrating global cues, including perspective convergence, occlusion ordering, and relative object sizes, that must propagate across the full $14\!\times\!14$ patch grid.  The 2--3 layer delay reflects this additional computational cost.

\textbf{Both degrade at layer 12.}  The final transformer block specialises for ImageNet class discrimination, discarding geometric information.  Boundary AP drops 26\% (0.833 to 0.617); depth MAE rises 13\% (0.0875 to 0.0992).

This progression, local structure before global spatial layout with both discarded at the classification head, parallels the classical visual cortex hierarchy (V1/V2 edges $\to$ V4/IT depth and shape).  The ViT appears to discover an analogous organisation through gradient descent on classification data alone.

\section{Causal analysis}
\label{sec:causal}

Probing establishes that depth information is \emph{present}.  Two causal interventions test whether it is \emph{structured} and \emph{actively maintained}.

\subsection{Probe direction ablation}

\begin{figure}[t]
\centering
\includegraphics[width=0.8\linewidth]{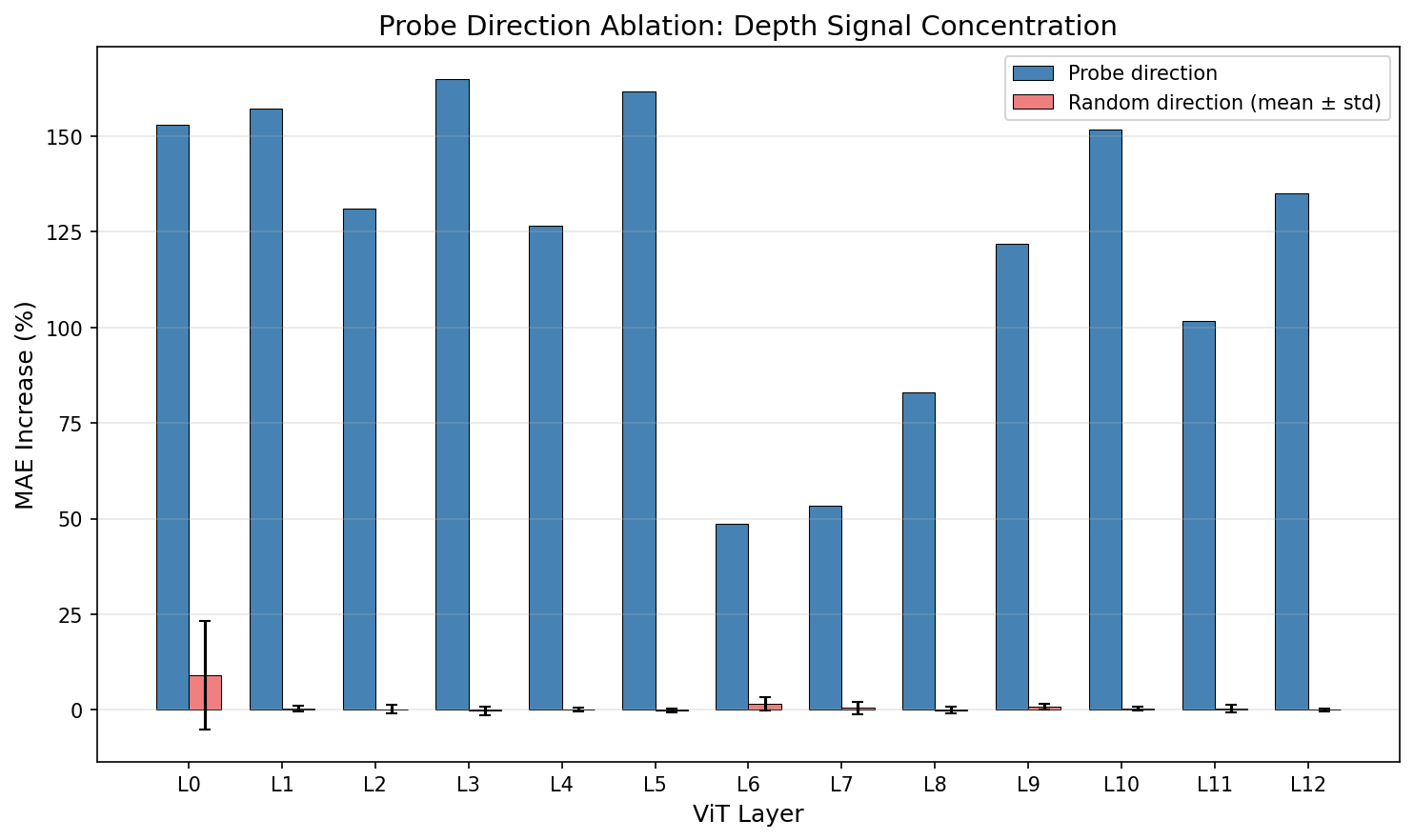}
\caption{MAE increase (\%) when ablating the probe's weight direction (blue) vs.\ a random direction (red).  The probe direction is load-bearing at every layer; random directions are not.}
\label{fig:ablation_gap}
\end{figure}

\begin{table}[t]
\centering
\caption{Ablation results.  Probe-direction ablation increases MAE by 49--165\%; random-direction ablation changes MAE by $<$1\%.}
\label{tab:ablation}
\begin{tabular}{@{}crrrr@{}}
\toprule
Layer & Orig.\ MAE & Ablated MAE & Gap\,\% & Random MAE \\
\midrule
0  & 0.1127 & 0.2850 & +153\% & $0.1228{\pm}0.016$ \\
5  & 0.0875 & 0.2291 & +162\% & $0.0874{\pm}0.001$ \\
6  & 0.0878 & 0.1305 & +49\%  & $0.0892{\pm}0.002$ \\
8  & 0.0875 & 0.1601 & +83\%  & $0.0874{\pm}0.001$ \\
12 & 0.0992 & 0.2331 & +135\% & $0.0992{\pm}0.000$ \\
\bottomrule
\end{tabular}
\end{table}

Across all 13 layers, ablating the probe direction increases MAE by \textbf{49--165\%}, while ablating a random direction changes it by $<$1\%.  This confirms depth is \textbf{concentrated in a specific linear direction}, not diffusely distributed.

The variation across layers tells its own story.  Layer 6 shows the smallest gap (+49\%) despite near-peak probe accuracy, suggesting the depth signal there may be partially distributed across more than one direction.  At layers 0, 5, and 12, the gap is larger (135--165\%), indicating tighter concentration in a single axis.  This is consistent with early and late layers having simpler depth encodings while mid-layers develop a richer subspace.

\begin{figure}[t]
\centering
\includegraphics[width=0.8\linewidth]{dose_response_layer08.png}
\caption{Dose-response at layer 8.  MAE increases smoothly as ablation strength $\alpha$ grows from 0 to 1.  No phase transition: the depth direction's contribution is graded.}
\label{fig:dose_response}
\end{figure}

The dose-response curve (Figure~\ref{fig:dose_response}) shows smooth, monotonic degradation with no sharp threshold, indicating a graded contribution of the depth direction to probe predictions.  This smooth scaling also addresses the ablation circularity concern: if the probe had simply memorised an arbitrary direction uncorrelated with the model's actual depth encoding, partially removing that direction would produce erratic rather than monotonic degradation, because the relationship between the ablated signal and the true encoding would be unpredictable.

\subsection{Targeted activation patching}

\begin{figure}[t]
\centering
\includegraphics[width=0.85\linewidth]{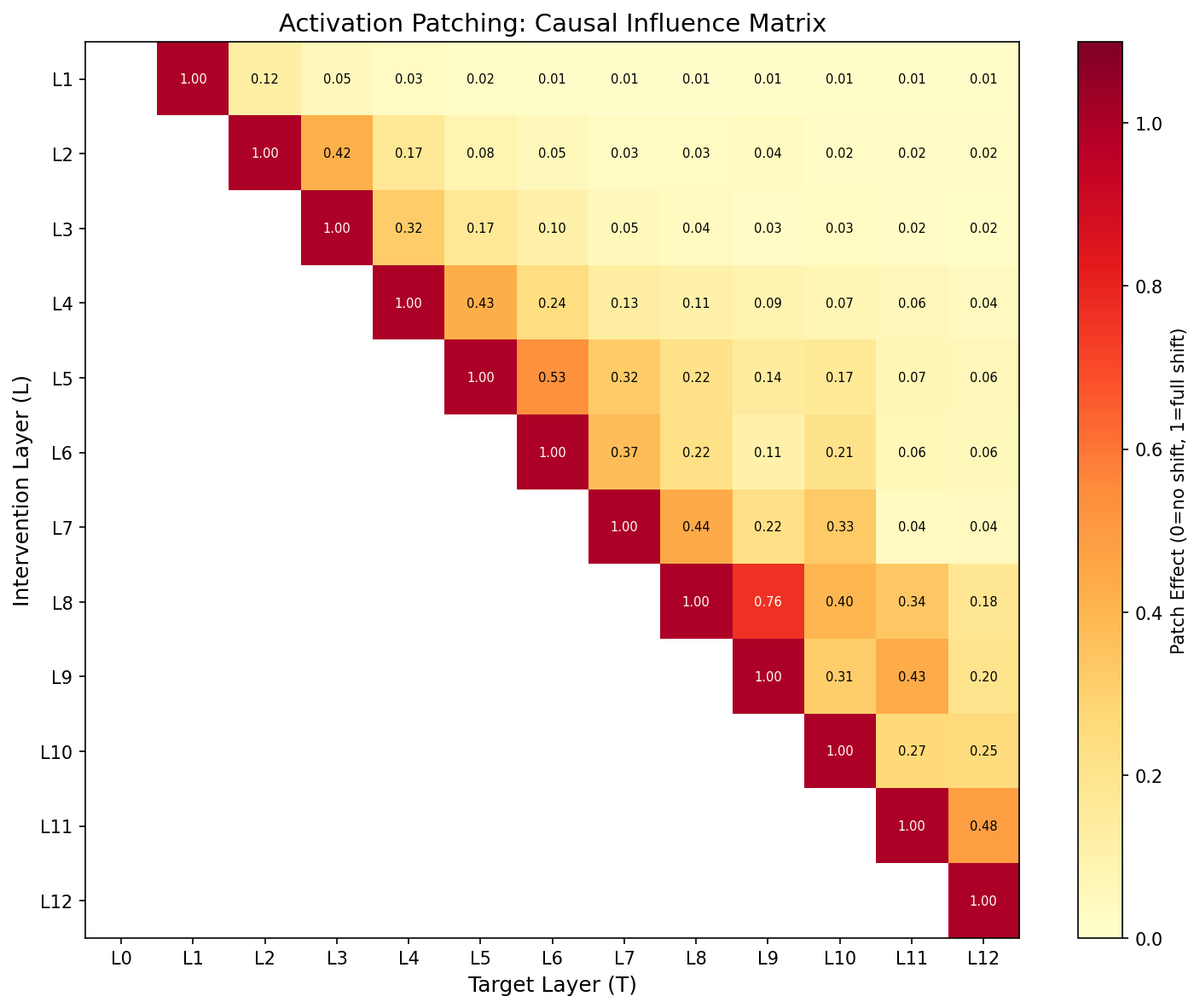}
\caption{Causal influence matrix.  Rows: intervention layer $L$.  Columns: measurement layer $T$.  Colour: mean patch effect across 20 image pairs.  Diagonal is 1.0 by construction; off-diagonal decay shows depth is re-derived, not passively carried.}
\label{fig:patching}
\end{figure}

\begin{table}[t]
\centering
\caption{Off-diagonal patch effects (mean across 20 pairs).  Mid-layer interventions persist farthest downstream.}
\label{tab:patching_offdiag}
\begin{tabular}{@{}crrrrr@{}}
\toprule
$L \backslash T$ & $T{=}L$ & $T{=}L{+}1$ & $T{=}L{+}2$ & $T{=}L{+}4$ & $T{=}12$ \\
\midrule
1  & 1.000 & 0.121 & 0.050 & 0.016 & 0.006 \\
5  & 1.000 & 0.535 & 0.323 & 0.166 & 0.057 \\
8  & 1.000 & 0.764 & 0.399 & 0.181 & 0.181 \\
11 & 1.000 & 0.483 & n/a   & n/a   & 0.483 \\
\bottomrule
\end{tabular}
\end{table}

The patching results support three conclusions:

\textbf{(1) Depth is partially re-derived at each layer.}  If depth were passively carried in the residual stream, swapping one direction would persist at near-1.0 through all downstream layers.  Instead, the effect decays: subsequent blocks partially overwrite the depth signal and re-compute it from the full 768-dimensional representation.

\textbf{(2) Mid-layer interventions persist most strongly.}  Patching at $L\!=\!8$ retains 76\% of the effect at $T\!=\!9$; patching at $L\!=\!1$ retains only 12\% at $T\!=\!2$.  Early-layer depth directions are fragile statistical correlations easily overwritten by subsequent attention blocks.  Mid-layer directions are stable geometric features.

\textbf{(3) The peak layer has the longest causal reach.}  At $L\!=\!8$, the effect at $T\!=\!12$ is still 0.181 (18\% of the depth prediction four layers downstream).  At $L\!=\!1$, it has decayed to 0.6\%.  Mid-layer depth representations are not only more accurate but also more causally robust.

\section{Discussion}

\subsection{Why mid-layer representations encode depth best}

Early layers (0--3) build local texture and edge representations.  Depth requires global context, including perspective convergence, occlusion ordering, and relative object sizes, that must propagate across the full $14\!\times\!14$ grid.  By layers 5--8, sufficient cross-patch communication has occurred for depth to become linearly decodable.  After layer 8, representations specialise for classification, discarding spatial detail not directly useful for distinguishing among 21,000 classes.

\subsection{Linear probes outperforming MLPs}

The linear probe outperforming the MLP on depth (0.0875 vs.\ 0.0920) despite $257\times$ fewer parameters indicates depth sits in a linearly separable subspace.  The MLP's extra capacity fits noise rather than useful nonlinear structure.  This aligns with the ablation finding: if depth were encoded nonlinearly across interacting dimensions, a linear probe would underperform and ablating one direction would not produce such dramatic effects.

\subsection{From correlation to causation}

The progression from probing (``depth is present'') to ablation (``the depth direction is load-bearing'') to targeted patching (``depth is actively re-derived'') builds a layered causal argument.  The ablation is the strongest single result: 165\% MAE increase from removing one direction out of 768, versus $<$1\% for any other.  The random-direction control rules out circularity.  The patching decay adds nuance: the ViT does not passively relay depth through the residual stream but partially re-derives it at each layer, with peak stability at the layers where probing accuracy is highest.

\subsection{The role of positional encodings}

The random-weight baseline's non-trivial depth performance (MAE $\approx$ 0.107) confirms that positional encodings contribute a floor of depth decodability, as discussed in Section 5.  The pretrained model builds on this floor with learned scene-specific features, achieving a 23\% reduction in MAE.  An important open question is whether replacing the standard learned positional encodings with alternatives (e.g., sinusoidal, rotary, or no positional encoding) would change the random baseline's floor or shift the layerwise profile of the pretrained model.  If the floor drops substantially without positional encodings, it would confirm that the random baseline's performance is primarily positional rather than architectural.

\section{Practical implications}

\textbf{Transfer learning.}  Mid-layer features (L5--8) should be preferred over the final layer for geometric downstream tasks such as monocular depth estimation, edge detection, or semantic segmentation.  A frozen mid-layer representation paired with a lightweight linear head could deliver strong geometric performance without fine-tuning later transformer blocks, saving substantial compute.  This is especially relevant for resource-constrained deployment, where freezing most of the model and training only a small head on top of a mid-layer is far cheaper than full fine-tuning.

\textbf{Model compression.}  Depth is concentrated in one direction out of 768.  When pruning or quantising a ViT for deployment in systems requiring spatial understanding (e.g., robotics perception, augmented reality), dimensions orthogonal to the depth direction can be compressed with minimal impact on geometric performance.  Conversely, pruning along the depth direction would be catastrophic.  The probing framework provides a principled way to identify load-bearing dimensions before making irreversible compression decisions.

\textbf{Failure diagnosis.}  If a ViT-based system produces incorrect depth predictions or misses object boundaries in a particular scene, layerwise probing can localise the failure.  By probing each layer for the relevant property on the failing input, a practitioner can determine whether the model failed to encode the property at the expected layer or whether the information was present at mid-layers but degraded before reaching the task head.  This narrows the debugging problem from a black-box failure to a layer-specific one.

\textbf{Architecture design.}  The spatial hierarchy motivates branching off mid-layer features for geometric heads, analogous to Feature Pyramid Networks in CNN architectures.  Our results provide specific empirical guidance for which layers to tap: L5--6 for structural features, L7--8 for spatial layout.  The finding that the final layer is actively hostile to geometric information suggests that multi-task architectures routing all tasks through the last layer leave geometric performance on the table.

\textbf{Concept auditing.}  The ablation methodology generalises beyond depth.  For any property of interest, one can train a linear probe, extract the encoding direction, ablate it, and measure downstream impact.  In safety-critical applications, this provides a tool for verifying that a model encodes properties it should rely on (e.g., lane markings for autonomous driving) or for detecting reliance on properties it should not encode (e.g., demographic attributes in a fairness-sensitive system).

\textbf{Pretraining strategy comparison.}  Our probing pipeline provides a standardised protocol for comparing how different training objectives organise spatial information.  Running the same layerwise probes on DINO, MAE, or DINOv2 would reveal whether self-supervised objectives preserve geometric information into later layers, shift the spatial hierarchy, or produce qualitatively different encoding structures.  This would directly inform practitioners choosing among pretraining strategies for geometric downstream tasks.

\section{Limitations}

\textbf{(1) Dataset scale and domain.}  NYU Depth V2 has 795 training images, all indoor.  BSDS500 has 200.  Depth results may not generalise to outdoor scenes where depth ranges and cues differ qualitatively.

\textbf{(2) Patch-level granularity.}  Probes operate at $16\!\times\!16$ resolution, losing fine-grained depth structure and collapsing the distinction between patches that barely clip a boundary and those centred on a strong edge.

\textbf{(3) Single model and training objective.}  We probe only ViT-B/16 with supervised ImageNet-21k weights.  Self-supervised models (DINO, MAE) or different scales (ViT-S, ViT-L) may produce different layerwise profiles.

\textbf{(4) Single-direction ablation.}  If depth occupies a low-dimensional subspace rather than a single axis, our ablation underestimates the total signal.  Multi-dimensional erasure (e.g., LEACE~\citep{belrose2023leace}) would be more thorough.

\textbf{(5) Boundary label noise.}  BSDS500 annotators frequently disagree, and majority-vote consensus leaves ambiguous patches noisy, acting as a ceiling on probe performance.

\textbf{(6) Ablation circularity.}  Probe-direction ablation by construction removes the direction the probe reads.  The random-direction control makes the result non-trivial, but the experiment proves the direction is load-bearing for the probe, not that the model uses depth for classification.  A fully causal claim would require intervening on classification logits.

\section{Conclusion}

We presented a systematic probing and causal intervention study of structural and spatial encoding in ViT-B/16, yielding five findings:

\begin{enumerate}[noitemsep,topsep=2pt,leftmargin=*]
    \item \textbf{Implicit encoding.}  Linear probes achieve AP = 0.833 for boundaries and MAE = 0.0875 for depth, both well above random baselines, confirming a classification-only ViT encodes both properties without supervision.

    \item \textbf{A spatial hierarchy.}  Boundaries peak at layers 5--6, depth at layer 8.  Local structure before global layout, paralleling the V1/V2 $\to$ V4/IT progression in the visual cortex.

    \item \textbf{Linear accessibility.}  Linear probes match or outperform MLPs, and ablation confirms depth is concentrated in a single linear direction (165\% MAE increase vs.\ $<$1\% for random directions).

    \item \textbf{Active maintenance.}  Targeted activation patching shows depth is partially re-derived at each layer: mid-layer interventions retain 76\% one layer downstream, while early-layer interventions decay to $<$5\% within two layers.

    \item \textbf{Final-layer degradation.}  Both properties collapse at layer 12, confirming the classification head discards spatial information.  Practitioners should tap mid-layer features for geometric tasks.
\end{enumerate}

These results demonstrate that a classification-trained ViT develops an actively maintained spatial hierarchy.  Rather than a passive residual-stream artefact, it is a causally structured encoding that the model re-derives and stabilises across its processing stack.

\end{document}